\documentclass[pdflatex,sn-mathphys]{sn-jnl}% Math and Physical Sciences Reference Style
\usepackage{caption}
\usepackage{subcaption}
\jyear{2022}%

\theoremstyle{thmstyleone}%
\theoremstyle{thmstyletwo}%

\theoremstyle{thmstylethree}%

\begin{document}

\title[The Platform for non-metallic pipes defects recognition.]
{The Platform for non-metallic pipes defects recognition. Design and Implementation.}


\author*[1]{\fnm{Fabio} \sur{Cacciatori}}\email{cacciatori@illogic.xyz}

\author[2]{\fnm{Sergei} \sur{Nikolaev}}\email{s.nikolaev@cyberphysics.xyz}

\author[1]{\fnm{Dmitrii} \sur{Grigorev}}\email{dm.srg.gr@gmail.com}

\affil*[1]{\orgdiv{R\&D}, \orgname{Illogic S.r.l.}, \orgaddress{\street{Street}, \city{Turin}, \postcode{100190}, \state{State}, \country{Itlay}}}

\affil[2]{\orgdiv{R\&D}, \orgname{Cyberphysics LLC}, \orgaddress{\street{Street}, \city{Moscow}, \postcode{10587}, \state{State}, \country{Russia}}}


\abstract{This paper describes a prototype software and hardware platform to provide support to field operators during the inspection of surface defects of non-metallic pipes.
Inspection is carried out by video filming defects created on the same surface in real-time using a "smart" helmet device and other mobile devices. The work focuses on the detection and recognition of the defects which appears as colored iridescence of reflected light caused by the diffraction effect arising from the presence of internal stresses in the inspected material.
The platform allows you to carry out preliminary analysis directly on the device in offline mode, and, if a connection to the network is established, the received data is transmitted to the server for post-processing to extract information about possible defects that were not detected at the previous stage.
The paper presents a description of the stages of design, formal description, and implementation details of the platform. It also provides descriptions of the models used to recognize defects and examples of the result of the work.
}




\maketitle

\section{Introduction}\label{sec1}
 
 Artificial vision is used for many industrial applications such as electronics component manufacturing\cite{bib1}, quality textile production \cite{bib2, bib3}, metal product finishing \cite{bib4}, glass manufacturing \cite{bib5}, printing products \cite{bib6}, and many others \cite{bib7}. However, when the inspection process requires a highly specific solution that rules out commercially available systems, manual inspection remains the most usual option. One example of a process usually performed by skilled workers is surface quality inspection. There are many reasons why these tasks prove very complex. First, many defects are visible only under certain lighting conditions and at a close distance. Second, the human operators have implicit limits: work turnover, low repeatability, and oversights of the defects or missing parts \cite{bib5}.

Finally, the inspection task must be carried out in the cycle time established for a specific production process, which tends to be short. These three reasons can have an impact on the production line and influence delivery time, production costs, and customer satisfaction. In particular, pipeline inspection is a part of pipeline integrity management for keeping the pipeline in good condition. The rules governing inspection are the pipeline safety regulations. In most cases, the pipeline is inspected regularly. The pipeline safety regulations require that the operator shall ensure that a pipeline is maintained in an efficient state, in efficient working order, and in good repair.

The pipeline inspection includes external inspection and, sometimes, internal inspection. Industrial pipeline systems are commonly used to transport oil, gas, and petrochemical products (e.g., corrosive substances). In-service inspection is required to avoid catastrophic failures and to guarantee the safe operation of pipelines \cite{bib8, bib9, bib10, bib11, bib12}. A defect is considered an elementary form of failure in pipes that could fail a safety system\cite{bib11}. 

A pipeline malfunctioning due to defects could lead to a reduction in or loss of profits in the oil, gas, and petrochemical industries. We refer to reports on some well-known failures, such as the leaked oil pipeline of the trans-Alaska pipeline system \cite{bib13} and the corroded gas pipeline in Guadalajara City in Mexico \cite{bib14}. Monitoring of defects (i.e., corrosion) in inaccessible regions, such as the interface between a pipe and the pipe supports \cite{bib15, bib16}, is sometimes infeasible by conventional non-destructive testing (NDT) methods \cite{bib11, bib17, bib18}. In these inaccessible areas, defects may develop rapidly and cause sudden failure \cite{bib19, bib20}.

The original technology is based on optical analysis of the material by processing images obtained by a reflected light color pattern and analyzing them. The structure of the color pattern corresponds to the structural changes in the inspected material and provides insights into potential destructive consequences of the defect’s development over time. In particular, the visual inspection is done by applying a film-based grating onto an inspected surface of a non-metallic pipe, resulting in a micro pattern of grooves. Such patterns produce colored iridescence of reflected light caused by diffraction effects on the mentioned microgrooves.

The main aim of this paper is to present the architecture that we are developing considering all the sub-systems involved in the architecture itself.

We would like to underline that, since the platform is in continuous development, we consider the possibility of slightly changes in the architecture and it's sub-systems.

A high-level representation of the architecture is reported in the figure below. The idea is an application with a GUI from which the operator can interact with some parameters (flattener filter dimension, video recording, etc.) that represents the core client application which contains video and images aquisition, preprocessing and defect detection parts using machine learning algorithms. Then the server-side where other machine learning algorithms are applied to detect defects, storage and database where the results are saved.
\begin{figure}[h]%
\centering
\includegraphics[width=0.8\textwidth]{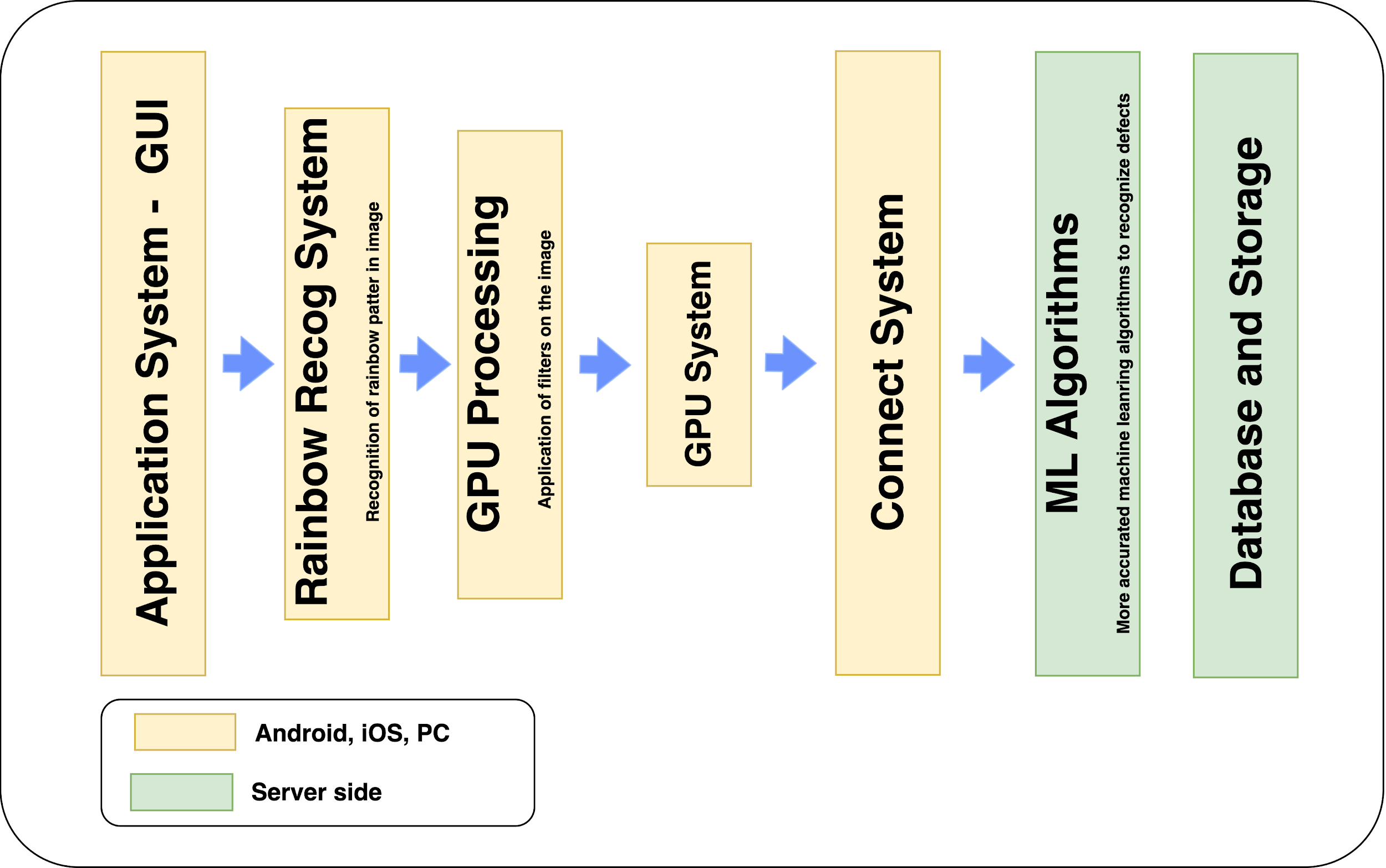}
\caption{Entire system architecture}\label{fig1}
\end{figure}

The entire architecture system can be divided into six different components.

\textit{ Application System}: this is the system where the GUI is launched. The operator can set all the parameters of the application, record a video of the inspection, etc.

\textit{Rainbow Recognition System}: it represents the implementation of the machine learning algorithms that have to check the presence of rainbow patterns in the image. It will give an advertisement to the operator that is recording the video.

\textit{GPU System}: it belongs to the functions that apply filters on the image (flattener filter) and the defect analyzer algorithm.

\textit{ML System}: is the core of the machine learning algorithms. It contains all the functions to detect and identify a defect. 

We have to highlight the fact that, during the implementation of the intelligent engine, we decided to implement the rainbow recognition system on the mobile device and the defect recognition system on the server-side only. Even using reduced network results (such as the one of rainbow recognition) the detection results were very poor, dropping down the performances of the entire application and having a high-power consumption.

\textit{Connect system}: it represents the connection layer between the annotated videos and the system where the whole information is stored.
DB and Storage System: it regards the database where the inspection information is stored and the storage to save the recorded inspection videos.

\section{Platform design}\label{sec2}

In this section, we would like to introduce the pipeline of the software. The figure below shows the functionality of the application.

\begin{figure}[h]%
\centering
\includegraphics[width=1\textwidth]{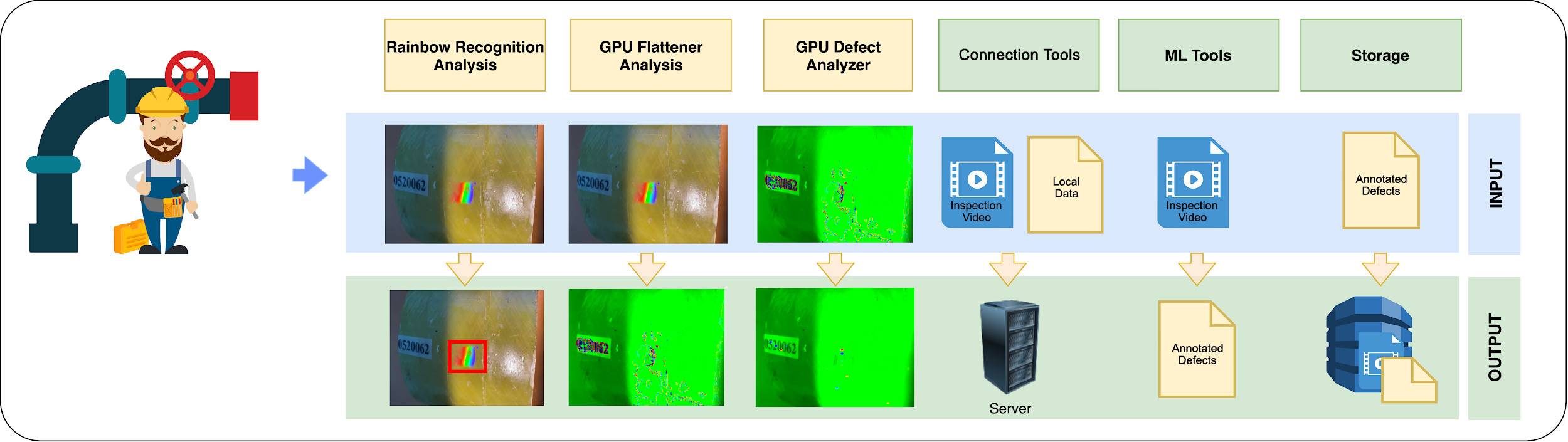}
\caption{Entire system architecture}\label{fig2}
\end{figure}
As shown in Fig. 2 software pipeline can be divided into 6 blocks:

1. Rainbow Recognition Analysis: the first block checks the presence of the rainbow color pattern in the image. If the rainbow is present and visible, an icon will notify it. It is useful to understand if the operator is recording a good inspection quality video.

2. GPU Flattener Analysis: this second block computes the flattener filter on the acquired image. The output, according to some parameters, is the flattened image. This part runs on the GPU of the device.

3. GPU Defect Analyzer: the third block, namely the GPU defect analyzer, is related to the application of the defect analyzer algorithm, to the flattened image computed at the previous stage.

4. Connection Tools: this block is related to the database and storage connection to save the data (video and annotations) after the inspection. The database is useful to retrieve information about a particular inspection, while the storage is used to save the recorded inspection video.

5. ML Tools (server): this block applies machine learning algorithms to an image to detect the defect type. We have implemented the ability to choose the most suitable machine learning solution to detect defects.

6. Storage: it represents the database and storage system.

%The main idea of the project is to perform architecture according to some conditions. %%
One of the critical requirements is the ability to perform inspection when the Internet connection is not available (i.e. a pipe in a desert or on an oil platform). For this reason, we create an application that works as a stand-alone and when the internet connection is available, it uploads all the data to the database and to the storage system.

The developed application is a cross-platform that run on PC, Android, or iOS-based systems. In the case of a PC application, the user have the possibility to download a video from the platform and run a complete inspection using the 3 main blocks – Rainbow Analysis, GPU Flattener Analysis, GPU Defect Analyzer, described in the previous paragraph. Concerning the MLTools, it use the machine learning algorithms on the server-side.
\subsection{Inspection Analysis Workflow}
The manual analysis of a previously saved inspection is always performed by an expert operator on a host that has dedicated GPU computing capabilities. To start analysis of an inspection the operator should open the “Inspections Library” and select a particular inspection and download it with all related data (such as rgb video, depth video and metadata on defect annotations). 
The operator can view the video playing on the screen and scroll through the various frames using the timeline at the top of the UI. For each frame the application shows the defect annotation (both manual or automatic).

Once the expert operator has evaluated the presence of a defect on a given frame, the operator will have the possibility to tag this defect on the current frame, choosing it from the defect classes previously set in the platform. 
Along with the defect type associated with the frame gets a screenshot of the current view as well as all the algorithm parameters. This will allow, when playing back the video, to restore the algorithm parameters used when the defect was saved for the first time. The operator has an option to delete a previously saved defect whether added by an operator or automatically.

Once the analysis is completed, the operator can upload the new defects list to the backend. As an option, the operator can restore the original defects list (undoing all the changes) by downloading the same inspection again.

\subsection{Intelligent engine}\label{subsec2}

The rainbow detection system is the first machine learning part of the entire architecture. It gives feedback to the operator that is capturing an inspection video, showing when the video is good or not. The idea is depicted in the figure below.

\begin{figure}[h]%
\centering
\includegraphics[width=0.9\textwidth]{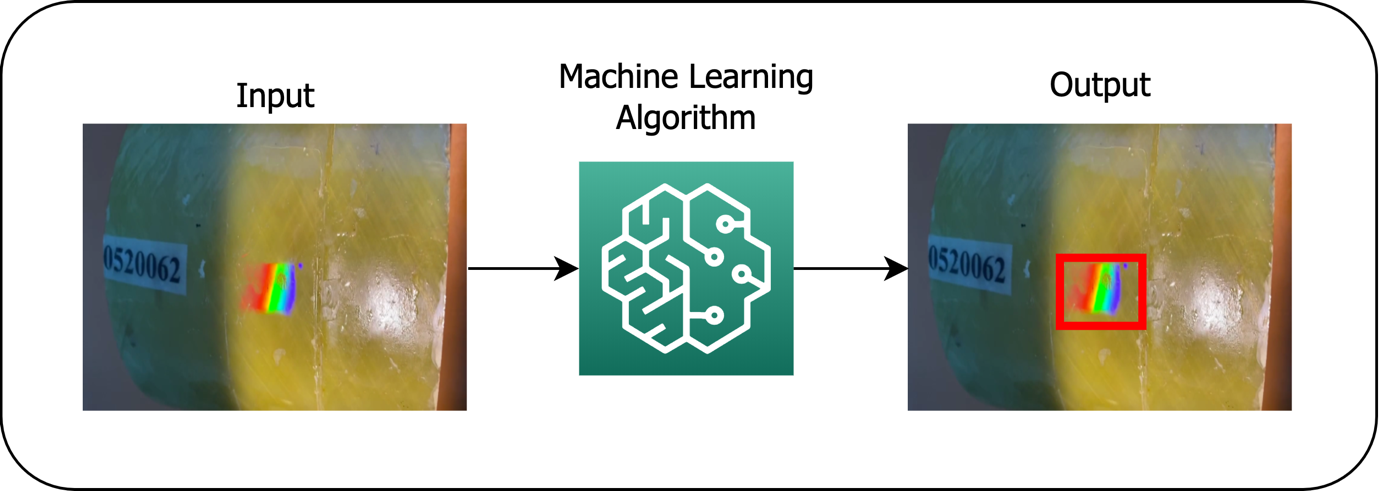}
\caption{Simplified idea of using Intelligent Engine}\label{fig3}
\end{figure}

The acquired frame is passed to the machine learning algorithm to identify the presence of the rainbow color pattern. 

A system icon will notify if the rainbow is visible, using a predefined system threshold. This is a helpful tool for the on-field operator

\subsection{Defect recognition}\label{subsubsec2}

The defect recognition system works similarly concerning rainbow detection.
The goal of this algorithm is to identify a particular defect and classifying it accordingly to one of the different classes. The image, a single frame of the video, is passed to the pre-trained deep learning model algorithm (CNN). The algorithm finds the most similar one, using different parameters, and it gives as output the type of defect and the accuracy of the recognition.
This part will be available on the server-side. The description of the entire method is depicted in the figure below.

\begin{figure}[h]%
\centering
\includegraphics[width=0.8\textwidth]{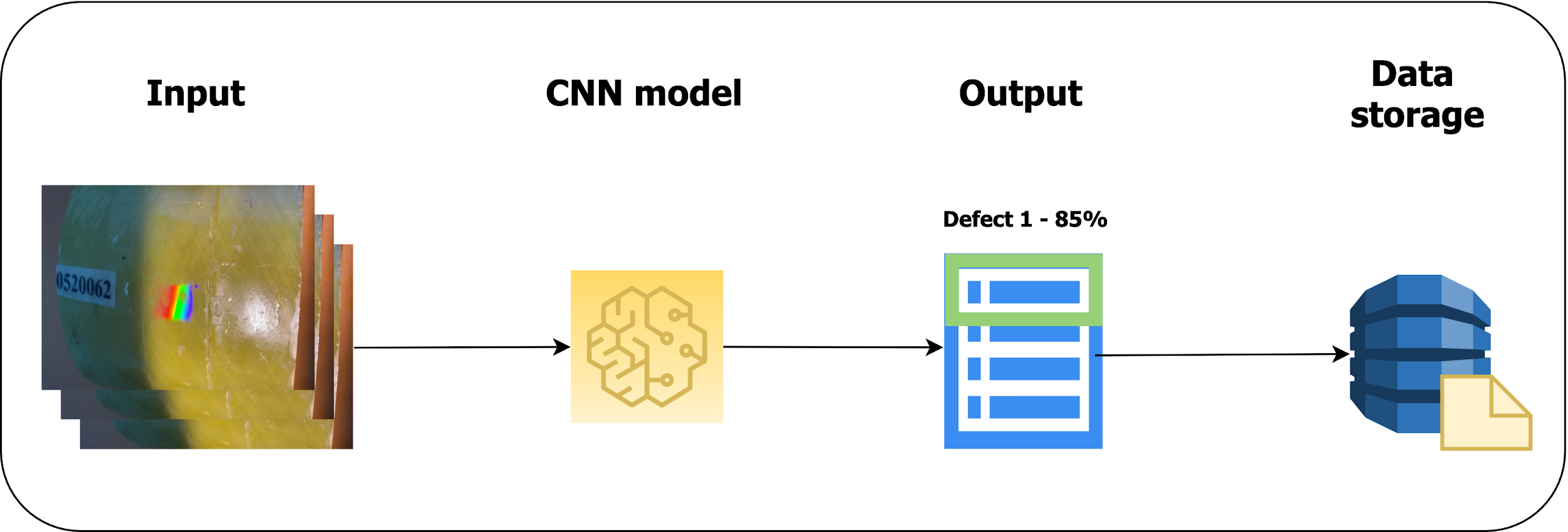}
\caption{Defect recognintion part}\label{fig4}
\end{figure}

\subsection{Machine learning back end}\label{subsubsec2}

The machine learning back-end of the web application has the main functionality of retraining the defect recognition algorithm. Since the number of images corresponding to different defects can be low, we decided to add a functionality that, using the information from an expert operator, allows to retrain the algorithm to improve the accuracy of the entire system. 

The operator, using the web application, will be able to create datasets, through a wizard, and retrain the network with new data. Since the algorithm needs labels and the position of the corresponding defect, There will be provided a tool to annotate the images and perform augmentation if necessary. The functionality of the platform also allows to store previously retrained models that implemented in a sort of backup system.

The entire procedure is shown in figure 5

\begin{figure}[h]%
\centering
\includegraphics[width=1.0\textwidth]{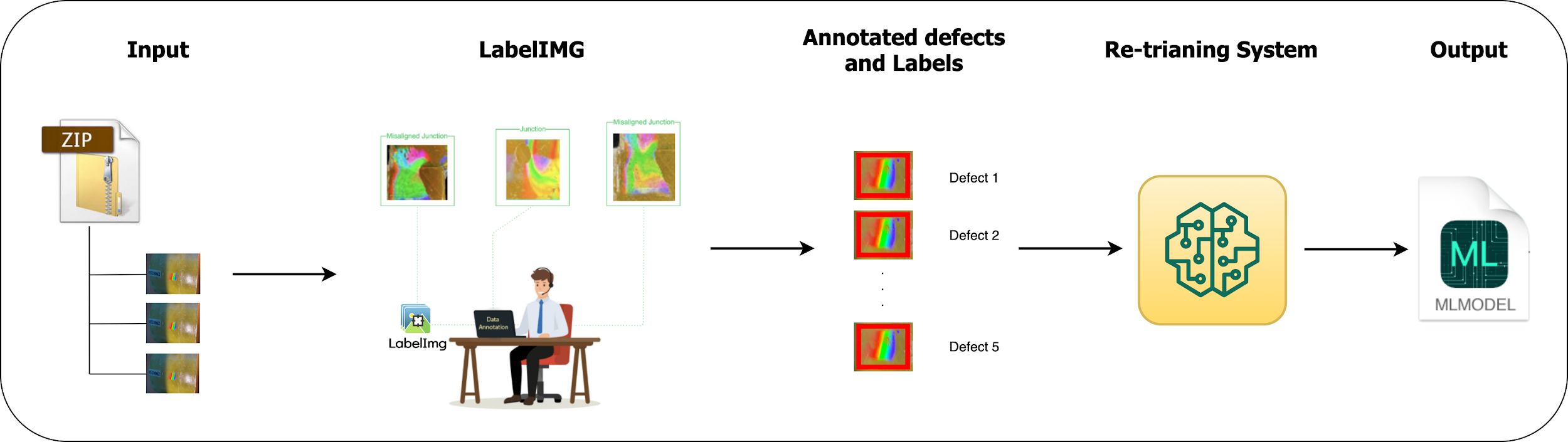}
\caption{Machine learning back-end pipeline}\label{fig5}
\end{figure}

\subsection{Web application}\label{subsubsec2}
The web application, as described previously, contains functionality:

Manage Users: this part is related to the management of the users registered in the system. The administrator will be able to register a new user, modify passwords, change roles, etc.

Statistics: it regards the possibility to have information from the database about top defects in the last 3 months or percentage per month of the number of inspections with defects over 1 year.

Semi-automatic defect annotation: the web application will have the possibility to get a video form the repository and shows all the set of related annotations. A simple GUI shows video and annotations.

Library: the library of annotated videos. Once a video is selected, the annotation and video itself are shown using the Semi-automatic defect annotation page.

ML system: this part of the web application focuses on the machine learning parts that can be supervised by the administrator: i.e. changing algorithm, retraining phase, machine learning active jobs, etc.

Tags: this page provide functionalities to create tags for a single inspection.

\section{Platform implementation}\label{sec3}

The application itself is developed to perform an intelligent visual inspection of plastic pipes. 
Also, it is worth noting that all further actions are motivated by the speed of implementation in order to obtain a minimum viable product. The choice of machine learning models, tuning options, and the final set of target metrics are taken to the final stage of development. Just like working with datasets.
In the current work, there are presented first approaches, which made it possible to build the entire process from receiving video inspections to post-processing and storing the results in the system.

The application is consist of three main parts:

1. Data Management System:
It represents the database where the data are stored (annotations and inspection videos);
For the implementation of the data management system, we decided to implement MongoDB to store inspection data. Another two databases are used by the system: The SQL database to store user information and the Redis database is used by the ml information system to store the tasks that have to be launched. 

2. Intelligent Engine: it represents the machine learning part of the system, considering both client and server-side;

3. Web application: it will focus on the management of the entire system, allowing the system admin to make queries, retrain the machine learning algorithms, manage the video library, etc.

The Intelligent engine covers different parts of the application system. It can be split into 2 main parts. The first one is related to rainbow recognition. It is a sub-system that runs on the client-side, helping the field operator that is capturing an inspection video. It gives an alert when a rainbow color pattern is recognized, reporting that a good inspection video is captured.

The second subsystem is the most completed one. The sub-systems that run on the client are less accurate, although faster since the client has less computational power than the server. The machine learning backend has the most complex algorithms to reach the best accuracy detection. Furthermore, it can be retrained (using the web application) and be constantly updated with new images fed into the algorithm.

For the implementation of the Intelligent engine we used TensorFlow\cite{tensorflow2015-whitepaper} and angular.
The CNNs chosen for the client-side and the server-side are different. This is done for two main reasons: first, the client-side has less computational power than the server; second, rainbow recognition (client-side) is a simpler problem for the defect recognition part (server-side).

\subsection{Server-Side}\label{subsubsec3}
At the beginning of development was decided to use the Faster R-CNN \cite{bib21} as the backbone model, by its computational efficiency (integrating the different training stages), low test time, and acceptable performance (mAP) on many tasks. 
Nevertheless, assuming the emergence of more efficient models the platform provides the functionality of implementing them in the pipeline.
The only stand-alone portion of the network left in Fast R-CNN was the region proposal algorithm. Both R-CNN and Fast R-CNN use the CPU-based region proposal algorithms, eg the Selective search algorithm which takes around 2 seconds per image and runs on CPU. The Faster R-CNN fixes this by using another convolutional network to generate the region proposals. This not only brings down the region proposal time from 2s to 10ms per image but also allows the region proposal stage to share layers with the following detection stages, causing an overall improvement in feature representation.

\subsection{Client-Side}\label{subsubsec3}

For the client-side (the rainbow recognition task), the main requirement is that the model should run on a mobile device and cannot be too complex since we want a real-time detection system. For this reason, we explored some “mobile” versions of CNN networks, such as\cite{bib22, bib23, bib24}:	SSD MobileNet and Tiny Yolo. First network is composed of two parts: SSD \cite{bib25} and MobileNet \cite{bib26}.
Tiny YOLO architecture was presented by the same authors of YOLO \cite{bib22,bib27}. The Tiny YOLO architecture is faster than its larger previous versions, achieving upwards of 244 FPS on a single GPU. The small model size (50MB), a reduced number of parameters, and fast inference speed make the Tiny-YOLO models more suited for embedded devices such as the Raspberry Pi, Google Coral, and NVIDIA Jetson Nano.

\section{Rainbow defect detection}\label{sec4}

Regarding the first sub-system, as discussed above, it was decided to use the Tiny Yolo v2 as backbone network for recognition of rainbow color patterns on mobile devices. We did different experiments to prove the effectiveness of our approach. 
First one,  using extracted images from the video recorded during inspection, with data augmentation (rotation, shifting, brightness, etc.) and the second, using extracted around 200 images from acquired video and other images. 

The annotation of the images was done using a LabelIMG tool \cite{bib28} in PASCAL Voc format. At the next stage of development, it is supposed to implement our own instrumentation to create custom data sets focused directly on various types of defects by the wizard procedure.

In this way, each image was associated with an XML file which contain all the information about the position of the bounding box that is used during the training of the network.
For Tiny Yolo, was used the DarkFlow implementation \cite{trieu2018darkflow}. 
Using the first tiny YOLO network (the one trained with 20 images) we noticed that the accuracy was very low and sometimes the rainbow pattern was not properly identified (see figures below).

\begin{figure}[h]
     \centering
     \begin{subfigure}[b]{0.4\textwidth}
         \centering
         \includegraphics[width=\textwidth]{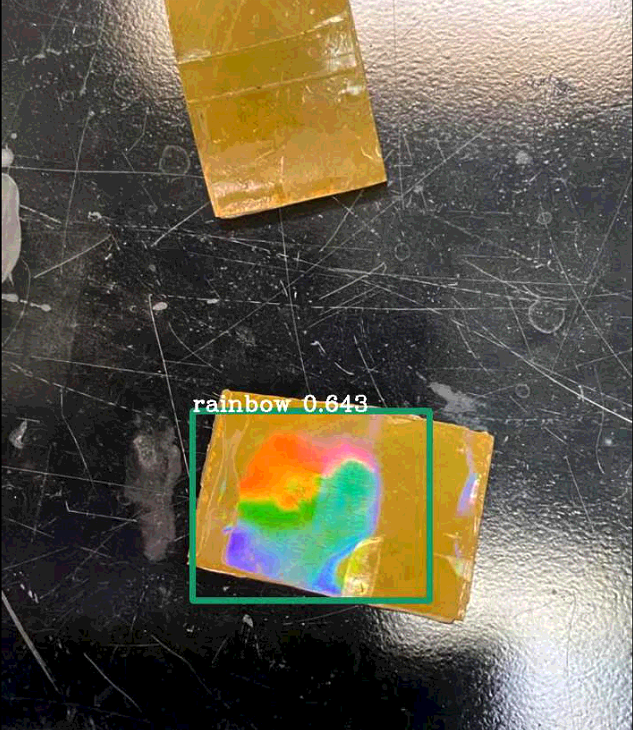}
         \caption{}
         \label{fig:y equals x}
     \end{subfigure}
     \begin{subfigure}[b]{0.4\textwidth}
         \centering
         \includegraphics[width=\textwidth]{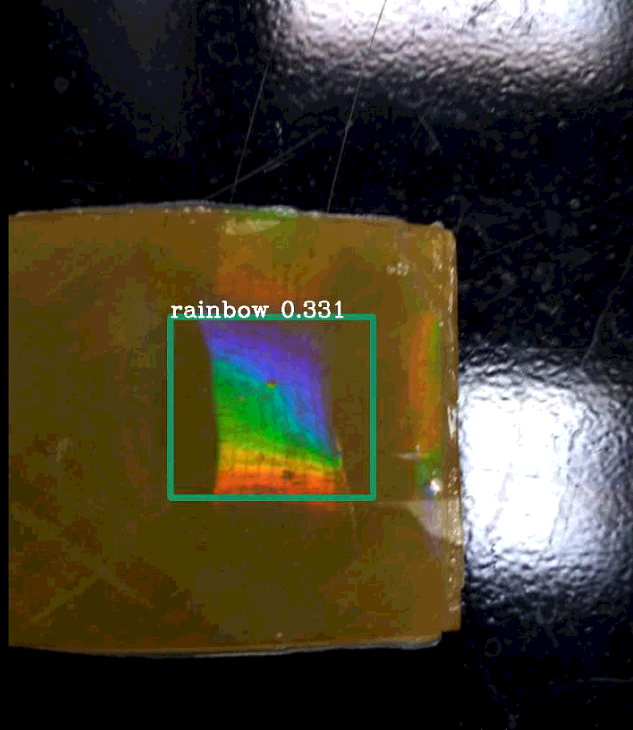}
         \caption{}
         \label{fig:five over x}
     \end{subfigure}
        \caption{Rainbow detection examples}
        \label{fig:Rainbow recognintion examples}
\end{figure}

Also the results using the second network are much better and the accuracy has been improved. The comparison was made using the same samples to show the effectiveness of this method.
The trained network was integrated inside the client application and tested on different cases. We decided to show an icon inside the app so when the field operator captures a video, an alert is shown to highlight the quality of the captured video.

\section{Defect recognition back-end}\label{sec5}

Regarding the back-end, we focused on the Faster R-CNN network, since this part should run on the server-side using the entire computational power of the machine. First of all, we would like to introduce the experiments that we did and then explain how we integrate the network in the back-end.
We ran four different experiments comparing both models( YOLO and Faster R-CNN):

1. Experiments with YOLOv2 network on flag\cite{flag} dataset.

2. Experiments with Faster R-CNN network on flag dataset.

3. Experiments with YOLOv2 network on rainbow defects dataset.

4. Experiments with Faster R-CNN network on rainbow defects dataset.

In the first two experiments, we decided to use a different datasets to test the effectiveness of our network in the detection of high saturated objects with different shapes. And, at the initial stage, we had to work with very few training samples with different defect types. Then we make a dataset with 600 original images and train both models. 
The results we get from experiments with the model used YOLO as the backbone are quite poor in terms of accuracy ($\sim$ 0.2). On the other hand, the other model results have higher accuracy ($\sim$ 0.8 - 0.9) and precision. 

In the third and fourth experiments, we used a small custom dataset of images with two classes: "Junction" and "Misaligned Junction",

First model (Faster R-CNN) is able to recognize the differences between the junction and misaligned junction class with good accuracy. The second (YOLO), instead, demonstrated low accuracy (less than 0.5). For this reason, in the end, we decided to move to the First (Faster R-CNN). For the final model, we use data augmentation to increase the number of images and we selected 648 images for training (324 per class) and 72 for testing (36 per class). As the overall result we add an option to select predefined model in the web application. 

\subsection{Retraining system}\label{subsec5}

The retraining system focuses on the functionality to train your own machine learning algorithm on a custom dataset. We have done a quick approbation test with selected 22 images split into 18 images for training (9 per class) and 4 images for a test (2 per class). The example results are reported in figures below.

\begin{figure}[h]
     \centering
     \begin{subfigure}[b]{0.4\textwidth}
         \centering
         \includegraphics[width=\textwidth]{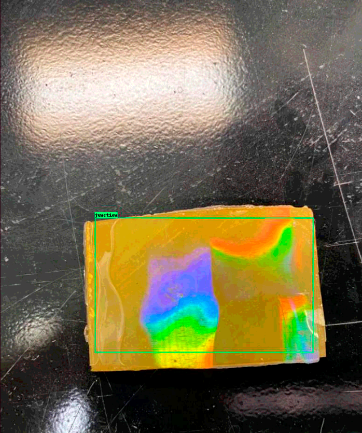}
         \caption{}
         \label{fig:y equals x}
     \end{subfigure}
     \begin{subfigure}[b]{0.4\textwidth}
         \centering
         \includegraphics[width=\textwidth]{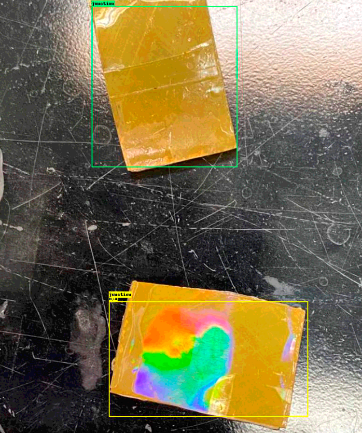}
         \caption{}
         \label{fig:five over x}
     \end{subfigure}
        \caption{Rainbow recognintion examples}
        \label{fig:Rainbow recognintion examples}
\end{figure}

We implement a backend procedure that passes a dataset to the system, it can start in a subprocess the retraining of the network. The subprocess is mandatory since we don’t want to block the web application when retraining is launched. In this way, a thread is called to run the entire process and, if new tasks are coming, they will be put in a queue list. The retraining procedure is done using TensorFlow Object Detection API. 
In the automatic script, we also inserted the possibility to download the model if it is not present in the server. The model is automatically extracted and ready to use. After this step, we 
have to modify the configuration file of the pre-trained model, according to our classes, steps, batch size, etc. To train the model, we run another script of Tensorflow.

1)	Jobs queue.
To implement the jobs queue system, we used a RQ service. RQ (Redis Queue) is a Python library for queueing jobs and processing them in the background with workers. It is backed by Redis and it is designed to have a low barrier to entry and can be integrated easily into any web stack. First, run a Redis server. You can use an existing one. To put jobs on queues, just define your blocking function (in our case the long processing function of retraining or video analysis) and then just call it using the enqueuer function of RQ.
We created a web page where it is possible to manage queues and delete them.

2)	Microservices Architecture.
The server side of the application was built by implementing the Microservices Architecture pattern. 
The application's backend is composed of several microservices:

•	cronjobs

•	ms-auth: to authenticates users and authorizes their access to some particular resources. It depends on the Mysql database in which users and roles are stored.

•	ms-entrypoint: this is the microservice that manages inspections, defects, and the analysis data.

•	ms-ml: to manage the machine learning part of the application. In particular, it controls the network management, the dataset management, the machine learning analysis of an inspection video, etc.

•	Nginx: to create a reverse proxy to map all the microservices addresses to a single point of access.
Also to get the server up and running, we use docker-compose. 

3)	Routing Architecture.
Routing or router in web development is a mechanism where HTTP requests are routed to the code that handles them. To put it simply, in the Router you determine what should happen when a user visits a certain page. We decided to use NGINX as a routing system.

\subsection{Web application}\label{subsec5}

The web application is fully implemented using Angular and Material design. The figures below show different pages of the developed web application.

\begin{figure}[h]%
\centering
\includegraphics[width=1\textwidth]{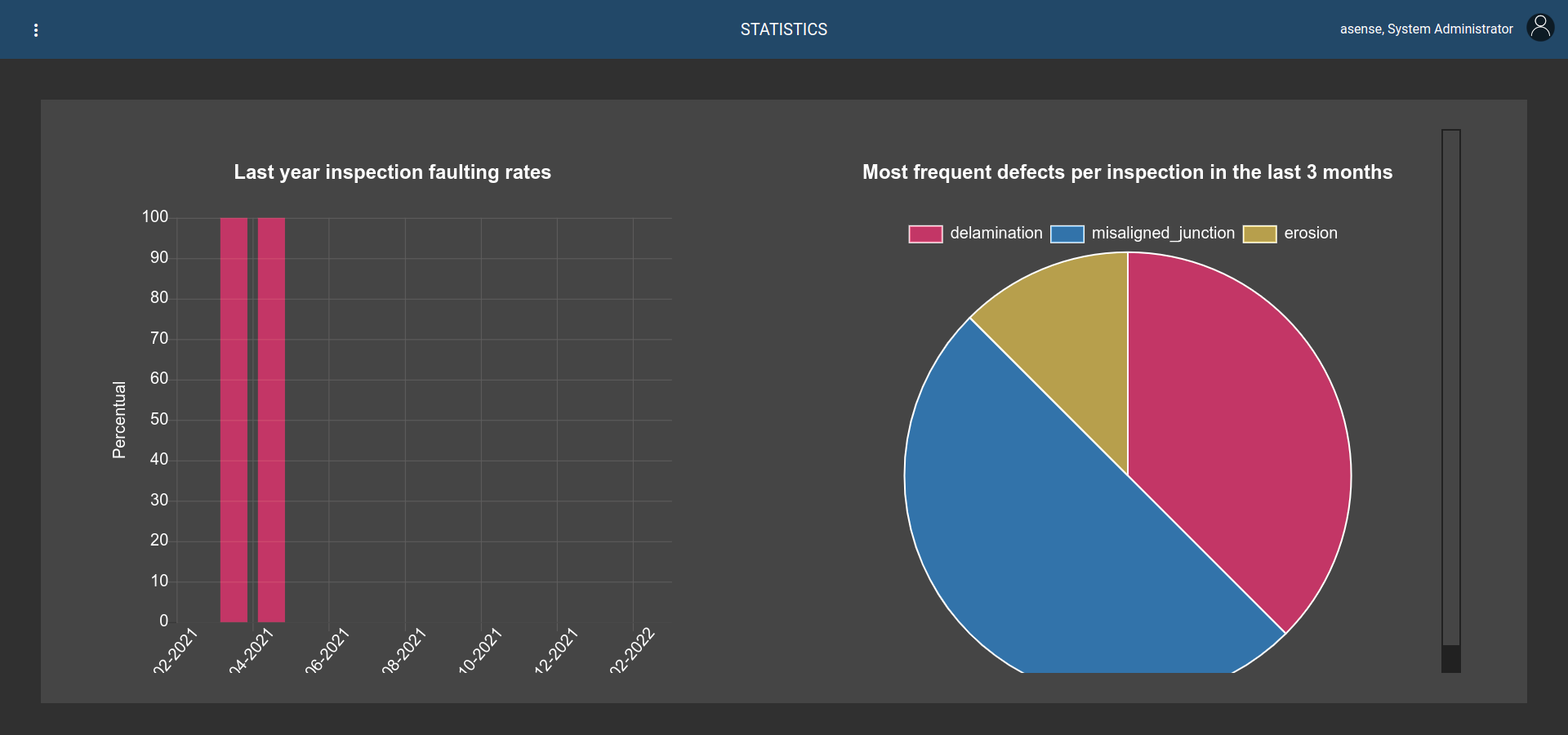}
\caption{Statistics page}\label{fig10}
\end{figure}

If a video is not locked (the ml analysis is complete), it is possible to click on it to show a
timeline of the video.

%Fig. 10. Statistics page  
\begin{figure}[h]%
\centering
\includegraphics[width=1\textwidth]{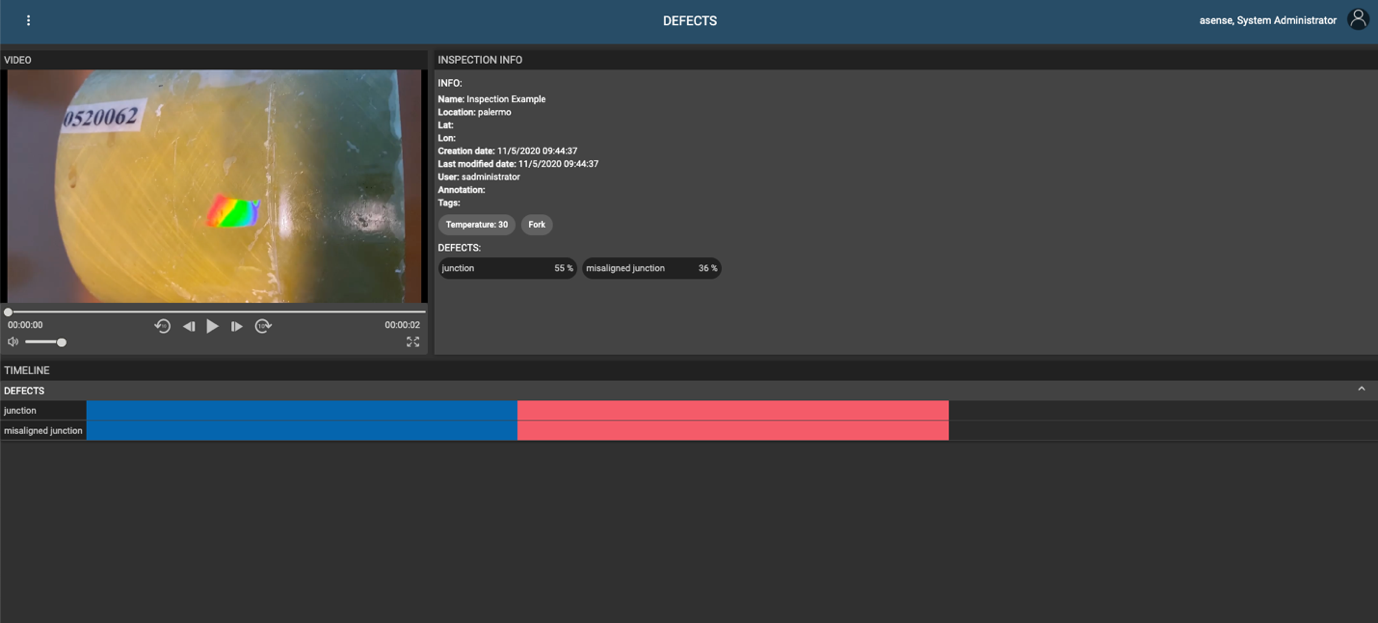}
\caption{Defects page with video timeline}\label{fig10}
\end{figure}
%Fig. 10. Defects page with video timeline
On the top left, the recorded video is shown. On the right, information about the video and the found defects are drawn. Furthermore, close to the name of the defect, a percentage gives the accuracy of detection. 
Login is controlled by an SQL database containing the username and password of a user. The access is granted by inserting correct values and using a web token.

%%\section{Figures}\label{sec6}

\section{Conclusion}\label{sec6}

In this work, we have proposed a platform designed from the ground up specifically for the detection and recognition of non-metallic pipes defects. 
The paper also presents the design process and provides specific approaches and solutions for the implementation of all parts of the platform. Our main aim is to make efficient use of scarce resources available on embedded platforms, compared to full-fledged deep learning workstations. 

The application of the developed system on the client-side hardware exemplifies real-time portable embedded solutions.

Even though the main goal was to run the application on portable devices, we separate some parts of machine learning algorithms on the server-side to post-process acquired video.

The developed platform will allow performing early and effective inspections of non-metallic pipes defects, which might lead to significant savings and minimize emergency situations.


\bibliography{sn-bibliography}% common bib file


\begin{thebibliography}{31}
\ifx \bisbn   \undefined \def \bisbn  #1{ISBN #1}\fi
\ifx \binits  \undefined \def \binits#1{#1}\fi
\ifx \bauthor  \undefined \def \bauthor#1{#1}\fi
\ifx \batitle  \undefined \def \batitle#1{#1}\fi
\ifx \bjtitle  \undefined \def \bjtitle#1{#1}\fi
\ifx \bvolume  \undefined \def \bvolume#1{\textbf{#1}}\fi
\ifx \byear  \undefined \def \byear#1{#1}\fi
\ifx \bissue  \undefined \def \bissue#1{#1}\fi
\ifx \bfpage  \undefined \def \bfpage#1{#1}\fi
\ifx \blpage  \undefined \def \blpage #1{#1}\fi
\ifx \burl  \undefined \def \burl#1{\textsf{#1}}\fi
\ifx \doiurl  \undefined \def \doiurl#1{\url{https://doi.org/#1}}\fi
\ifx \betal  \undefined \def \betal{\textit{et al.}}\fi
\ifx \binstitute  \undefined \def \binstitute#1{#1}\fi
\ifx \binstitutionaled  \undefined \def \binstitutionaled#1{#1}\fi
\ifx \bctitle  \undefined \def \bctitle#1{#1}\fi
\ifx \beditor  \undefined \def \beditor#1{#1}\fi
\ifx \bpublisher  \undefined \def \bpublisher#1{#1}\fi
\ifx \bbtitle  \undefined \def \bbtitle#1{#1}\fi
\ifx \bedition  \undefined \def \bedition#1{#1}\fi
\ifx \bseriesno  \undefined \def \bseriesno#1{#1}\fi
\ifx \blocation  \undefined \def \blocation#1{#1}\fi
\ifx \bsertitle  \undefined \def \bsertitle#1{#1}\fi
\ifx \bsnm \undefined \def \bsnm#1{#1}\fi
\ifx \bsuffix \undefined \def \bsuffix#1{#1}\fi
\ifx \bparticle \undefined \def \bparticle#1{#1}\fi
\ifx \barticle \undefined \def \barticle#1{#1}\fi
\bibcommenthead
\ifx \bconfdate \undefined \def \bconfdate #1{#1}\fi
\ifx \botherref \undefined \def \botherref #1{#1}\fi
\ifx \url \undefined \def \url#1{\textsf{#1}}\fi
\ifx \bchapter \undefined \def \bchapter#1{#1}\fi
\ifx \bbook \undefined \def \bbook#1{#1}\fi
\ifx \bcomment \undefined \def \bcomment#1{#1}\fi
\ifx \oauthor \undefined \def \oauthor#1{#1}\fi
\ifx \citeauthoryear \undefined \def \citeauthoryear#1{#1}\fi
\ifx \endbibitem  \undefined \def \endbibitem {}\fi
\ifx \bconflocation  \undefined \def \bconflocation#1{#1}\fi
\ifx \arxivurl  \undefined \def \arxivurl#1{\textsf{#1}}\fi
\csname PreBibitemsHook\endcsname

%%% 1
\bibitem{bib1}
\begin{barticle}
\bauthor{\bsnm{Park}, \binits{Y.}},
\bauthor{\bsnm{Kweon}, \binits{I.S.}}:
\batitle{Ambiguous surface defect image classification of amoled displays in
  smartphones}.
\bjtitle{IEEE Transactions on Industrial Informatics}
\bvolume{12}(\bissue{2}),
\bfpage{597}--\blpage{607}
(\byear{2016}).
\doiurl{10.1109/TII.2016.2522191}
\end{barticle}
\endbibitem

%%% 2
\bibitem{bib2}
\begin{barticle}
\bauthor{\bsnm{Bahlmann}, \binits{C.}},
\bauthor{\bsnm{Heidemann}, \binits{G.}},
\bauthor{\bsnm{Ritter}, \binits{H.}}:
\batitle{Artificial neural networks for automated quality control of textile
  seams}.
\bjtitle{Pattern Recognition}
\bvolume{32}(\bissue{6}),
\bfpage{1049}--\blpage{1060}
(\byear{1999}).
\doiurl{10.1016/S0031-3203(98)00128-9}
\end{barticle}
\endbibitem

%%% 3
\bibitem{bib3}
\begin{barticle}
\bauthor{\bsnm{Kumar}, \binits{A.}}:
\batitle{Computer-vision-based fabric defect detection: A survey}.
\bjtitle{IEEE Transactions on Industrial Electronics}
\bvolume{55}(\bissue{1}),
\bfpage{348}--\blpage{363}
(\byear{2008}).
\doiurl{10.1109/TIE.1930.896476}
\end{barticle}
\endbibitem

%%% 4
\bibitem{bib4}
\begin{botherref}
\oauthor{\bsnm{Tucker}, \binits{J.}}:
Inside beverage can inspection: An application from start to finish.
Proceedings of the Vision Conference,
97--107
(1989)
\end{botherref}
\endbibitem

%%% 5
\bibitem{bib5}
\begin{barticle}
\bauthor{\bsnm{Peng}, \binits{X.}},
\bauthor{\bsnm{Chen}, \binits{Y.}},
\bauthor{\bsnm{Yu}, \binits{W.}},
\bauthor{\bsnm{Zhou}, \binits{Z.}},
\bauthor{\bsnm{Sun}, \binits{G.}}:
\batitle{An online defects inspection method for float glass fabrication based
  on machine vision}.
\bjtitle{The International Journal of Advanced Manufacturing Technology}
\bvolume{39}(\bissue{11}),
\bfpage{1180}--\blpage{1189}
(\byear{2008}).
\doiurl{10.1007/s00170-007-1302-7}
\end{barticle}
\endbibitem

%%% 6
\bibitem{bib6}
\begin{barticle}
\bauthor{\bsnm{Torres}, \binits{F.}},
\bauthor{\bsnm{Sebastian}, \binits{J.M.}},
\bauthor{\bsnm{Aracil}, \binits{R.}},
\bauthor{\bsnm{Jimenez}, \binits{L.M.}},
\bauthor{\bsnm{Reinoso}, \binits{O.}}:
\batitle{Automated real-time visual inspection system for high-resolution
  superimposed printings}.
\bjtitle{Image and Vision Computing}
\bvolume{16}(\bissue{12}),
\bfpage{947}--\blpage{958}
(\byear{1998}).
\doiurl{10.1016/S0262-8856(98)00059-6}
\end{barticle}
\endbibitem

%%% 7
\bibitem{bib7}
\begin{barticle}
\bauthor{\bsnm{Malamas}, \binits{E.N.}},
\bauthor{\bsnm{Petrakis}, \binits{E.G.M.}},
\bauthor{\bsnm{Zervakis}, \binits{M.}},
\bauthor{\bsnm{Petit}, \binits{L.}},
\bauthor{\bsnm{Legat}, \binits{J.-D.}}:
\batitle{A survey on industrial vision systems, applications and tools}.
\bjtitle{Image and Vision Computing}
\bvolume{21}(\bissue{2}),
\bfpage{171}--\blpage{188}
(\byear{2003}).
\doiurl{10.1016/S0262-8856(02)00152-X}
\end{barticle}
\endbibitem

%%% 8
\bibitem{bib8}
\begin{bbook}
\bauthor{\bsnm{Lin}, \binits{S.}},
\bauthor{\bsnm{Ito}, \binits{T.}},
\bauthor{\bsnm{Kawashima}, \binits{K.}},
\bauthor{\bsnm{Nagamizo}, \binits{H.}}:
In: \beditor{\bsnm{Thompson}, \binits{D.O.}},
\beditor{\bsnm{Chimenti}, \binits{D.E.}} (eds.)
\bbtitle{Finite Element Analysis of Multiple Wave Scattering from Defects
  within a Circular Pipe},
pp. \bfpage{79}--\blpage{85}.
\bpublisher{Springer},
\blocation{Boston, MA}
(\byear{1999}).
\doiurl{10.1007/978-1-4615-4791-4\textunderscore 9}.
\burl{https://doi.org/10.1007/978-1-4615-4791-4\textunderscore 9}
\end{bbook}
\endbibitem

%%% 9
\bibitem{bib9}
\begin{bbook}
\bauthor{\bsnm{Lowe}, \binits{M.}}:
In: \beditor{\bsnm{Thompson}, \binits{D.O.}},
\beditor{\bsnm{Chimenti}, \binits{D.E.}} (eds.)
\bbtitle{Characteristics of the Reflection of Lamb Waves from Defects in Plates
  and Pipes},
pp. \bfpage{113}--\blpage{120}.
\bpublisher{Springer},
\blocation{Boston, MA}
(\byear{1998}).
\doiurl{10.1007/978-1-4615-5339-7\textunderscore14}.
\burl{https://doi.org/10.1007/978-1-4615-5339-7\textunderscore14}
\end{bbook}
\endbibitem

%%% 10
\bibitem{bib10}
\begin{barticle}
\bauthor{\bsnm{Galdos}, \binits{A.}},
\bauthor{\bsnm{Okuda}, \binits{H.}},
\bauthor{\bsnm{Yagawa}, \binits{G.}}:
\batitle{Finite element simulation of ultrasonic wave propagation in pipe and
  pressure vessel walls}.
\bjtitle{Finite Elements in Analysis and Design}
\bvolume{7}(\bissue{1}),
\bfpage{1}--\blpage{13}
(\byear{1990}).
\doiurl{10.1016/0168-874X(90)90011-3}
\end{barticle}
\endbibitem

%%% 11
\bibitem{bib11}
\begin{bchapter}
\bauthor{\bsnm{Lee}, \binits{P.-H.}},
\bauthor{\bsnm{Yang}, \binits{S.-K.}}:
\bctitle{Defect inspection of complex structure in pipes by guided waves}.
In: \beditor{\bsnm{Wu}, \binits{T.-T.}},
\beditor{\bsnm{Ma}, \binits{C.-C.}} (eds.)
\bbtitle{IUTAM Symposium on Recent Advances of Acoustic Waves in Solids},
pp. \bfpage{389}--\blpage{395}.
\bpublisher{Springer},
\blocation{Dordrecht}
(\byear{2010})
\end{bchapter}
\endbibitem

%%% 12
\bibitem{bib12}
\begin{barticle}
\bauthor{\bsnm{Zhu}, \binits{W.}}:
\batitle{{An FEM Simulation for Guided Elastic Wave Generation and Reflection
  in Hollow Cylinders With Corrosion Defects}}.
\bjtitle{Journal of Pressure Vessel Technology}
\bvolume{124}(\bissue{1}),
\bfpage{108}--\blpage{117}
(\byear{2001})
{\href{https://arxiv.org/abs/https://asmedigitalcollection.asme.org/pressurevesseltech/article-pdf/124/1/108/5675159/108\_1.pdf}{{https://asmedigitalcollection.asme.org/pressurevesseltech/article-pdf/124/1/108/5675159/108\_1.pdf}}}.
\doiurl{10.1115/1.1428331}
\end{barticle}
\endbibitem

%%% 13
\bibitem{bib13}
\begin{botherref}
Oil Leak Is Latest Mishap for Alaska’s Troubled Pipelines.
Available online
  \url{https://www.propublica.org/article/oil-leak-is-latest-mishap-for-troubled-alaska-pipeline-system.}
(2002)
\end{botherref}
\endbibitem

%%% 14
\bibitem{bib14}
\begin{botherref}
Sewers explode in Guadalajara.
Available online
  \url{https://www.history.com/this-day-in-history/sewers-explode-in-guadalajara.}
(2002)
\end{botherref}
\endbibitem

%%% 15
\bibitem{bib15}
\begin{barticle}
\bauthor{\bsnm{Satyarnarayan}, \binits{L.}},
\bauthor{\bsnm{Chandrasekaran}, \binits{J.}},
\bauthor{\bsnm{Maxfield}, \binits{B.}},
\bauthor{\bsnm{Balasubramaniam}, \binits{K.}}:
\batitle{Circumferential higher order guided wave modes for the detection and
  sizing of cracks and pinholes in pipe support regions}.
\bjtitle{NDT \& E International}
\bvolume{41}(\bissue{1}),
\bfpage{32}--\blpage{43}
(\byear{2008}).
\doiurl{10.1016/j.ndteint.2007.07.004}
\end{barticle}
\endbibitem

%%% 16
\bibitem{bib16}
\begin{barticle}
\bauthor{\bsnm{Balasubramaniam}, \binits{K.}},
\bauthor{\bsnm{Chandrasekaran}, \binits{J.}},
\bauthor{\bsnm{Maxfield}, \binits{B.W.}},
\bauthor{\bsnm{Satyanarayan}, \binits{L.}}:
\batitle{Imaging hidden corrosion using ultrasonic non‐dispersive higher
  order guided wave modes}.
\bjtitle{AIP Conference Proceedings}
\bvolume{975}(\bissue{1}),
\bfpage{215}--\blpage{222}
(\byear{2008})
{\href{https://arxiv.org/abs/https://aip.scitation.org/doi/pdf/10.1063/1.2902661}{{https://aip.scitation.org/doi/pdf/10.1063/1.2902661}}}.
\doiurl{10.1063/1.2902661}
\end{barticle}
\endbibitem

%%% 17
\bibitem{bib17}
\begin{barticle}
\bauthor{\bsnm{Belanger}, \binits{P.}},
\bauthor{\bsnm{Cawley}, \binits{P.}}:
\batitle{Feasibility of low frequency straight-ray guided wave tomography}.
\bjtitle{NDT \& E International}
\bvolume{42}(\bissue{2}),
\bfpage{113}--\blpage{119}
(\byear{2009}).
\doiurl{10.1016/j.ndteint.2008.10.006}
\end{barticle}
\endbibitem

%%% 18
\bibitem{bib18}
\begin{barticle}
\bauthor{\bsnm{Yibo}, \binits{L.}},
\bauthor{\bsnm{Liying}, \binits{S.}},
\bauthor{\bsnm{Zhidong}, \binits{S.}},
\bauthor{\bsnm{Yuankai}, \binits{Z.}}:
\batitle{Study on energy attenuation of ultrasonic guided waves going through
  girth welds}.
\bjtitle{Ultrasonics}
\bvolume{44},
\bfpage{1111}--\blpage{1116}
(\byear{2006}).
\doiurl{10.1016/j.ultras.2006.05.108}.
\bcomment{Proceedings of Ultrasonics International (UI’05) and World Congress
  on Ultrasonics (WCU)}
\end{barticle}
\endbibitem

%%% 19
\bibitem{bib19}
\begin{bbook}
\bauthor{\bsnm{Su}, \binits{Z.}},
\bauthor{\bsnm{Ye}, \binits{L.}}:
\bbtitle{Lecture Notes in Applied and Computational Mechanics. Identification
  of Damage Using Lamb Waves: From Fundamentals to Applications}.
\bpublisher{Springer},
\blocation{London}
(\byear{2009})
\end{bbook}
\endbibitem

%%% 20
\bibitem{bib20}
\begin{botherref}
\oauthor{\bsnm{Shivaraj}, \binits{K.}},
\oauthor{\bsnm{Balasubramaniam}, \binits{K.}},
\oauthor{\bsnm{Krishnamurthy}, \binits{C.V.}},
\oauthor{\bsnm{Wadhwan}, \binits{R.}}:
{Ultrasonic Circumferential Guided Wave for Pitting-Type Corrosion Imaging at
  Inaccessible Pipe-Support Locations}.
Journal of Pressure Vessel Technology
\textbf{130}(2)
(2008)
{\href{https://arxiv.org/abs/https://asmedigitalcollection.asme.org/pressurevesseltech/article-pdf/130/2/021502/5534861/021502\_1.pdf}{{https://asmedigitalcollection.asme.org/pressurevesseltech/article-pdf/130/2/021502/5534861/021502\_1.pdf}}}.
\doiurl{10.1115/1.2892031}.
021502
\end{botherref}
\endbibitem

%%% 21
\bibitem{tensorflow2015-whitepaper}
\begin{botherref}
\oauthor{\bsnm{Abadi}, \binits{M.}},
\oauthor{\bsnm{Agarwal}, \binits{A.}},
\oauthor{\bsnm{Barham}, \binits{P.}},
\oauthor{\bsnm{Brevdo}, \binits{E.}},
\oauthor{\bsnm{Chen}, \binits{Z.}},
\oauthor{\bsnm{Citro}, \binits{C.}},
\oauthor{\bsnm{Corrado}, \binits{G.S.}},
\oauthor{\bsnm{Davis}, \binits{A.}},
\oauthor{\bsnm{Dean}, \binits{J.}},
\oauthor{\bsnm{Devin}, \binits{M.}},
\oauthor{\bsnm{Ghemawat}, \binits{S.}},
\oauthor{\bsnm{Goodfellow}, \binits{I.}},
\oauthor{\bsnm{Harp}, \binits{A.}},
\oauthor{\bsnm{Irving}, \binits{G.}},
\oauthor{\bsnm{Isard}, \binits{M.}},
\oauthor{\bsnm{Jia}, \binits{Y.}},
\oauthor{\bsnm{Jozefowicz}, \binits{R.}},
\oauthor{\bsnm{Kaiser}, \binits{L.}},
\oauthor{\bsnm{Kudlur}, \binits{M.}},
\oauthor{\bsnm{Levenberg}, \binits{J.}},
\oauthor{\bsnm{Man\'{e}}, \binits{D.}},
\oauthor{\bsnm{Monga}, \binits{R.}},
\oauthor{\bsnm{Moore}, \binits{S.}},
\oauthor{\bsnm{Murray}, \binits{D.}},
\oauthor{\bsnm{Olah}, \binits{C.}},
\oauthor{\bsnm{Schuster}, \binits{M.}},
\oauthor{\bsnm{Shlens}, \binits{J.}},
\oauthor{\bsnm{Steiner}, \binits{B.}},
\oauthor{\bsnm{Sutskever}, \binits{I.}},
\oauthor{\bsnm{Talwar}, \binits{K.}},
\oauthor{\bsnm{Tucker}, \binits{P.}},
\oauthor{\bsnm{Vanhoucke}, \binits{V.}},
\oauthor{\bsnm{Vasudevan}, \binits{V.}},
\oauthor{\bsnm{Vi\'{e}gas}, \binits{F.}},
\oauthor{\bsnm{Vinyals}, \binits{O.}},
\oauthor{\bsnm{Warden}, \binits{P.}},
\oauthor{\bsnm{Wattenberg}, \binits{M.}},
\oauthor{\bsnm{Wicke}, \binits{M.}},
\oauthor{\bsnm{Yu}, \binits{Y.}},
\oauthor{\bsnm{Zheng}, \binits{X.}}:
{TensorFlow}: Large-Scale Machine Learning on Heterogeneous Systems.
Software available from tensorflow.org
(2015).
\url{https://www.tensorflow.org/}
\end{botherref}
\endbibitem

%%% 22
\bibitem{bib21}
\begin{botherref}
\oauthor{\bsnm{Ren}, \binits{S.}},
\oauthor{\bsnm{He}, \binits{K.}},
\oauthor{\bsnm{Girshick}, \binits{R.}},
\oauthor{\bsnm{Sun}, \binits{J.}}:
Faster R-CNN: Towards Real-Time Object Detection with Region Proposal Networks.
arXiv
(2015).
\doiurl{10.48550/ARXIV.1506.01497}.
\url{https://arxiv.org/abs/1506.01497}
\end{botherref}
\endbibitem

%%% 23
\bibitem{bib22}
\begin{botherref}
\oauthor{\bsnm{Mehta}, \binits{R.}},
\oauthor{\bsnm{Ozturk}, \binits{C.}}:
Object detection at 200 Frames Per Second.
arXiv
(2018).
\doiurl{10.48550/ARXIV.1805.06361}.
\url{https://arxiv.org/abs/1805.06361}
\end{botherref}
\endbibitem

%%% 24
\bibitem{bib23}
\begin{botherref}
\oauthor{\bsnm{Redmon}, \binits{J.}},
\oauthor{\bsnm{Divvala}, \binits{S.}},
\oauthor{\bsnm{Girshick}, \binits{R.}},
\oauthor{\bsnm{Farhadi}, \binits{A.}}:
You Only Look Once: Unified, Real-Time Object Detection.
arXiv
(2015).
\doiurl{10.48550/ARXIV.1506.02640}.
\url{https://arxiv.org/abs/1506.02640}
\end{botherref}
\endbibitem

%%% 25
\bibitem{bib24}
\begin{botherref}
\oauthor{\bsnm{Redmon}, \binits{J.}},
\oauthor{\bsnm{Farhadi}, \binits{A.}}:
YOLO9000: Better, Faster, Stronger.
arXiv
(2016).
\doiurl{10.48550/ARXIV.1612.08242}.
\url{https://arxiv.org/abs/1612.08242}
\end{botherref}
\endbibitem

%%% 26
\bibitem{bib25}
\begin{bchapter}
\bauthor{\bsnm{Liu}, \binits{W.}},
\bauthor{\bsnm{Anguelov}, \binits{D.}},
\bauthor{\bsnm{Erhan}, \binits{D.}},
\bauthor{\bsnm{Szegedy}, \binits{C.}},
\bauthor{\bsnm{Reed}, \binits{S.}},
\bauthor{\bsnm{Fu}, \binits{C.-Y.}},
\bauthor{\bsnm{Berg}, \binits{A.C.}}:
\bctitle{Ssd: Single shot multibox detector}.
In: \beditor{\bsnm{Leibe}, \binits{B.}},
\beditor{\bsnm{Matas}, \binits{J.}},
\beditor{\bsnm{Sebe}, \binits{N.}},
\beditor{\bsnm{Welling}, \binits{M.}} (eds.)
\bbtitle{Computer Vision -- ECCV 2016},
pp. \bfpage{21}--\blpage{37}.
\bpublisher{Springer},
\blocation{Cham}
(\byear{2016})
\end{bchapter}
\endbibitem

%%% 27
\bibitem{bib26}
\begin{botherref}
\oauthor{\bsnm{Howard}, \binits{A.G.}},
\oauthor{\bsnm{Zhu}, \binits{M.}},
\oauthor{\bsnm{Chen}, \binits{B.}},
\oauthor{\bsnm{Kalenichenko}, \binits{D.}},
\oauthor{\bsnm{Wang}, \binits{W.}},
\oauthor{\bsnm{Weyand}, \binits{T.}},
\oauthor{\bsnm{Andreetto}, \binits{M.}},
\oauthor{\bsnm{Adam}, \binits{H.}}:
MobileNets: Efficient Convolutional Neural Networks for Mobile Vision
  Applications.
arXiv
(2017).
\doiurl{10.48550/ARXIV.1704.04861}.
\url{https://arxiv.org/abs/1704.04861}
\end{botherref}
\endbibitem

%%% 28
\bibitem{bib27}
\begin{botherref}
\oauthor{\bsnm{Redmon}, \binits{J.}}:
Darknet: Open Source Neural Networks in C.
\url{http://pjreddie.com/darknet/}
(2013--2016)
\end{botherref}
\endbibitem

%%% 29
\bibitem{bib28}
\begin{botherref}
\oauthor{\bsnm{Tzutalin}}:
LabelImg. Git code.
Available online \url{https://github.com/tzutalin/labelImg}
(2015)
\end{botherref}
\endbibitem

%%% 30
\bibitem{trieu2018darkflow}
\begin{botherref}
\oauthor{\bsnm{Trieu}, \binits{T.H.}}:
Darkflow.
GitHub Repository. Available online: https://github.com/thtrieu/darkflow
  (accessed on 14 February 2019)
(2018)
\end{botherref}
\endbibitem

%%% 31
\bibitem{flag}
\begin{botherref}
\oauthor{\bsnm{Dua}, \binits{D.}},
\oauthor{\bsnm{Graff}, \binits{C.}}:
{UCI} Machine Learning Repository
(2017).
\url{http://archive.ics.uci.edu/ml}
\end{botherref}
\endbibitem

\end{thebibliography}


\end{document}